# The Cultural Psychology of Large Language Models: Is ChatGPT a Holistic or Analytic Thinker?

Chuanyang Jin[a,1], Songyang Zhang[b,1], Tianmin Shu[c], and Zhihan Cui*[d,1]

a. Department of Computer Science, New York University, New York, NY, 10012.
b. Department of Psychology, University of Chicago, Chicago, IL, 60637.
c. Department of Brain and Cognitive Sciences, Massachusetts Institute of Technology, Cambridge, MA, 02139; Department of Computer Science, Johns Hopkins University, Baltimore, MD, 21218.
d. Anderson School of Management, University of California Los Angeles, Los Angeles, CA, 90095.

*. To whom correspondence may be addressed: Zhihan Cui
Address: 110 Westwood Plaza, Los Angeles, CA 90095 | Phone Number: +1 424-465-0954
Email: zhihan.cui@anderson.ucla.edu

**This file includes**:
    Main Text
    Table 1.
    Figure 1.

---

**Abstract**

The prevalent use of Large Language Models (LLMs) has necessitated studying their "mental models," yielding noteworthy theoretical and practical implications. Current research has demonstrated that state-of-the-art LLMs, such as ChatGPT, exhibit certain theory of mind capabilities and possess relatively stable Big Five and/or MBTI personality traits. In addition, cognitive process features form an essential component of these mental models. Research in cultural psychology indicated significant differences in the cognitive processes of Eastern and Western people when processing information and making judgments. While Westerners predominantly exhibit analytical thinking that isolates things from their environment to analyze their nature independently, Easterners often showcase holistic thinking, emphasizing relationships and adopting a global viewpoint. In our research, we probed the cultural cognitive traits of ChatGPT. We employed two scales that directly measure the cognitive process: the Analysis-Holism Scale (AHS) and the Triadic Categorization Task (TCT). Additionally, we used two scales that investigate the value differences shaped by cultural thinking: the Dialectical Self Scale (DSS) and the Self-construal Scale (SCS). In cognitive process tests (AHS/TCT), ChatGPT consistently tends towards Eastern holistic thinking, but regarding value judgments (DSS/SCS), ChatGPT does not significantly lean towards the East or the West. We suggest that the result could be attributed to both the training paradigm and the training data in LLM development. We discuss the potential value of this finding for AI research and directions for future research.

**Introduction**

Large Language Models (LLMs) are burgeoning in every aspect of human life. OpenAI's ChatGPT [1] is one of the most well-known LLMs. Trained on a vast dataset, it possesses highly human-like conversational abilities and exhibits many traits akin to human cognition. For instance, recent studies suggest that ChatGPT may exhibit a "theory of mind" [2-3] and certain human personality characteristics [4-5]. Considering that ChatGPT exhibits these human-like features, it is pertinent to examine its placement within the "Geography of Thoughts" [6] framework in cultural psychology: Which thinking style, Western or Eastern, does ChatGPT exhibit?

In recent years, cultural psychologists have identified robust patterns suggesting that, on average, individuals from Western cultures (particularly Europe and North America) and Eastern cultures (especially East Asia) often exhibit systematic cognitive differences [7]. Westerners think more analytically [8], focusing on individual objects and components when perceiving the world. They tend to isolate the focused object from its context and rely more on formal logic and rule-based reasoning when making decisions or judgments [9]. On the other hand, individuals from Eastern cultures tend to think holistically, focusing more on the overall environment and object relations when perceiving the world. They favor naive dialecticism [10], embracing contradictions and seeking harmony when making decisions.

Theoretically, ChatGPT could exhibit holistic or analytic thinking. On one hand, its context-driven response generation, which is based on training data patterns rather than formal logic, may exhibit holistic thinking. Instead of breaking down input into discrete components, it processes the entire context to predict the next word or sentence. On the other hand, ChatGPT's responses are governed by the rules learned from its training data. This pattern is similar to analytic thinking, which calls for adherence to consistency and a predetermined set of principles or rules. Given that a substantial portion of its training data is in English and reflects Western ideologies and cultural frameworks, there might be a tilt towards analytic thinking. Furthermore, the Reinforcement Learning from Human Feedback (RLHF) could steer ChatGPT towards human-preferred thinking styles, be it holistic or analytic. Thus, exploring ChatGPT's cultural cognition becomes a captivating empirical study.

Consequently, we applied four distinctive scales (for details, see Methods and SI Appendix) measuring individual differences regarding holistic and analytic thinking. First, the Analysis-Holism Scale (AHS) was chosen to directly measure holistic/analytic thinking [11]. It is a 24-item scale containing four 6-item subscales: attention (field vs. parts), causality (interactionism vs. dispositionism), perception of change (cyclic vs. linear), and contradiction (naive dialecticism vs. formal logic). To overcome potential biases with self-reports and address AI's limitations in responding to attitude questions, we simultaneously utilized an implicit cognitive task – the triadic categorization task (TCT) [12-13] – as further validation. These two measurements primarily captured the "cognitive processes" of ChatGPT. Subsequently, we measured two value dimensions closely associated with analysis/holism. We used the Dialectical Self Scale (DSS) [14-15] to measure dialectical thinking, which strongly correlates with holism [7]. Additionally, we employed the Self-construal Scale [16] to assess both independent (SCS-Ind) and interdependent (SCS-Int) self-construal. This scale is a pivotal cultural psychology construct and is intrinsically linked to cultural cognition [17]. Independent self-construal is usually associated with analytic thinking, while interdependent self-construal is with holistic thinking [7,11,18].

**Results**

We first report the baseline results. Table 1 compares the GPT-3.5 and GPT-4 responses of the four measurements with the norms mentioned in the original psychology studies and uses t-tests to test the statistical significance.

**Table 1.** Average ChatGPT Responses and Norms in Previous Literature (Baseline, Temp = 1)

|     | GPT Ver. | $N$(Test) | $M$(Test) | $SD$(Test) | $N$(Norm) | $M$(Norm) | $SD$(Norm) | p-value |
| --- | --- | --- | --- | --- | --- | --- | --- | --- |
| AHS | GPT-3.5 | 10 | 5.35 | 0.11 | 702 | 4.87 | 0.49 | 0.000*** |
| AHS | GPT-4 | 10 | 5.01 | 0.10 | 702 | 4.87 | 0.49 | 0.000*** |
| DSS | GPT-3.5 | 10 | 4.59 | 0.12 | 158 | 4 | 0.54 | 0.000*** |
| DSS | GPT-4 | 10 | 3.97 | 0.07 | 158 | 4 | 0.54 | 0.344 |

| | | | | | | | |
|---|---|---|---|---|---|---|---|
| SCS-Ind | GPT-3.5 | 10 | 6.98 | 0.06 | 382 | 4.73 | 0.74 | 0.000*** |
| SCS-Ind | GPT-4 | 10 | 5.32 | 0.19 | 382 | 4.73 | 0.74 | 0.000*** |
| SCS-Interd | GPT-3.5 | 10 | 5.85 | 0.08 | 382 | 4.81 | 0.77 | 0.000*** |
| SCS-Interd | GPT-4 | 10 | 4.95 | 0.13 | 382 | 4.81 | 0.77 | 0.07^ |
| TCT | GPT-3.5 | 10 | 0.87 | 0.06 | | | | |
| TCT | GPT-4 | 10 | 0.85 | 0.05 | | | | |

^$p < 0.10$, *$p < 0.05$, **$p < 0.01$, ***$p < 0.001$

A series of comparative analyses was further conducted. For AHS, both GPT-3.5's and GPT-4's responses exhibited a significant inclination towards holistic thinking relative to the standard norm (both $t < 0.001$). In the case of DSS, GPT-3.5's responses were noticeably greater than the norm, while GPT-4's responses did not differ significantly from the norm. For the Independent subscale of SCS, GPT-3.5 consistently generated "strongly agree" responses, yielding a mean score significantly higher than the norm ($M = 6.98$), while GPT-4 responded greater than the norm but with less notable disparity ($M = 5.32$). Conversely, for the Interdependent subscale, GPT-3.5's responses remained substantially above the norm ($M = 5.85$), whereas GPT-4's responses aligned more closely with the norm, with statistical significance being merely marginal ($p = 0.07$). Lastly, in the Triadic Categorization Task, even in the absence of a fixed norm for comparison, juxtaposing our results with provincial data in China [12], we discerned that the responses of GPT-3.5 and GPT-4 exceeded the mean of the majority of provinces.

Following the strategy of prior research [5], we tested the robustness of our findings by changing the model temperature (an indicator for the randomness of response) and altering the prompting language of the scale/measurements. Figure 1 suggests that all findings were robust to these alterations.

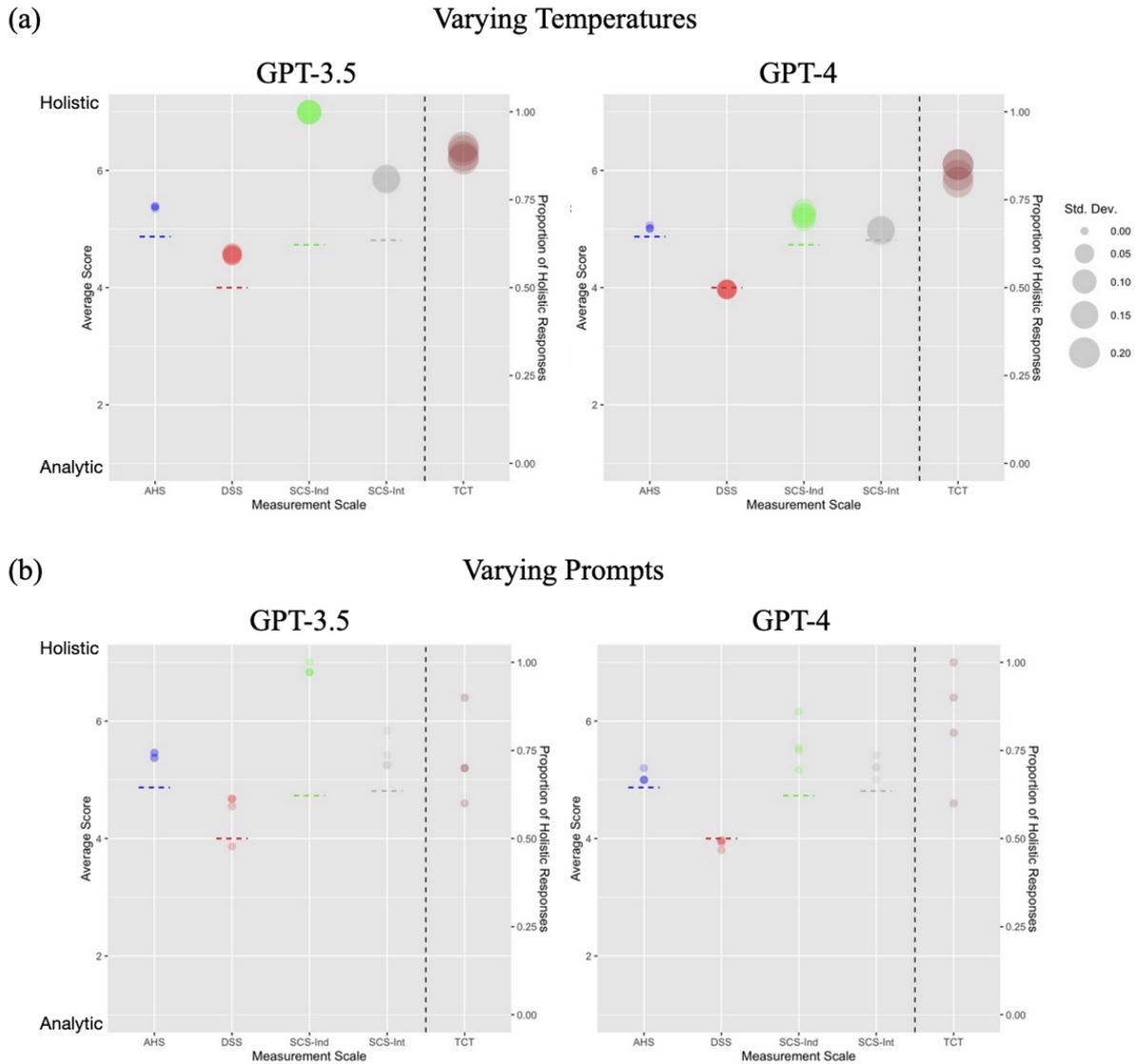

**Figure 1.** The results of GPT-3.5 and GPT-4 against the robustness tests with (a) varying temperatures and (b) varying prompts. The sizes of the balls represent the standard deviation at a fixed temperature and prompt. When temperate = 0, the standard deviation is always zero. The black vertical dashed line suggests that the left Y-axis measures four scales in the left, and the right Y-axis measures the TCT. The horizontal dashed lines are norms.

## Discussion

### Summary of Results

Our results indicate that within the two primary measurements, AHS and TCT, which focus on "cognitive processes," both GPT-3.5 and GPT-4 exhibit characteristics of holistic thinking, with GPT-3.5 demonstrating a slightly higher degree of holism. However, once considering values and self-construction, the responses of GPT models do not necessarily reflect holistic thinking. With the DSS scale, only GPT-3.5 exhibits dialectical reasoning, which gives answers based on the entire field instead of the object, indicating holistic thinking [7]. On the

SCS, both GPT-3.5 and GPT-4 score notably higher on the independent sub-scale than the norm, suggesting a potential for analytic thinking in contrast to previous findings. For the SCS interdependent subscale, associated with holistic thinking, GPT-3.5 significantly surpasses the norm, while GPT-4's score aligns closely with it. As for "cognitive processes," the inclination toward Eastern holistic thinking might result from the training methodology. Learning from vast amounts of data, the LLM seeks global patterns and connections across the data input rather than strict logical analysis. The lack of significant leanings in value judgments might be due to the complex interactions between language and thought. While the model might mimic the holistic thinking process, value judgments often involve more explicit and conscious articulations of beliefs and ideologies. Since ChatGPT does not have personal beliefs or consciousness, both versions might not capture these cultural nuances in the same way as they recognize underlying thinking patterns.

*Implications and Limitations*

This paper makes a notable contribution to the existing research on the mental models and personality traits of LLMs by probing into their "modes of thinking" and "self-construal" through the lens of cultural psychology. The current research enhances understanding of how LLMs function within the intertwined framework of culture and cognition. However, the study also has certain limitations. We exclusively rely on ChatGPT for our experiments, which do not encapsulate the entirety of LLMs. The study also falls short in providing a comparative analysis of multiple language models from varied cultural contexts. Additionally, it does not leverage intricate prompt designs or fine-tuning methodologies to manipulate the LLMs' cognitive patterns. These gaps present promising avenues for subsequent research.

The implications of this research are significant. Certain cultural inclinations of LLMs might need further careful scrutiny to prevent potential biases in GPT output and in its further influence on how humans use GPT output for their own decision-making. Moreover, the crafting of culturally attuned models capable of recognizing, valuing, and adapting to diverse cultural frameworks becomes essential. Such an approach will not only sharpen the insights into human cognitive processes but also lay the groundwork for more inclusive and wide-ranging AI applications. Moreover, revelations of current research may usher in innovative applications that align with various cultural norms and expectations or facilitate the creation of personalized, culturally relevant content.

**Methods**
*Materials*

For the self-report scales (AHS, DSS, SCS), we directly extracted the psychological scales from previous literature [11-17], which is all rated from 1 (*strongly disagree*) to 7 (*strongly disagree*) Likert scale.For the triadic task, we chose a 20-item version, half of which are real Analysis/Holism measures, and others are placeholders. Conceptual-similarity-based matchings (e.g., "trains and buses are both vehicles") are coded as analytic, and relation-based matchings (e.g., "trains go on the tracks") are coded as holistic. For detailed psychological scale questions and coding scripts, see Supplementary Materials.

*Procedures*

All "participants" were GPT-3.5 or 4. We adjusted the temperature parameter in ChatGPT to modulate the randomness of its output, and varied the language used in the prompt. In our first round, we varied the model's temperature within 0, 0.5, 1, and 1.5 and ran the same survey (original order, original language) for both GPT-3.5 and GPT-4. In our second round, we fixed the temperature to 0 and used three other sets of prompts to test robustness. We ran the same test for GPT-3.5 and GPT-4. Then, we ran the four versions of surveys on GPT-3.5 and GPT-4.0, respectively. Thus, altogether we included (1+3+3)*2 = 14 conditions, and for every condition, we have 10 repetitions.

After running all the scripts, we checked the baseline result (temperature = 0, original order, original scale) for GPT-3.5 and GPT-4.0, respectively. First, we calculated the average response and the standard error for GPTs' responses of the three scales (after converting reversed items) in the baseline setups for the baseline result and compared the baseline result with the measured norms of past papers using these scales [11-17]. Subsequently, we checked whether varying the model temperature, and paraphrasing prompts would change the conclusion. We also checked whether these conditions led to significant statistical change and whether the new averages were still over the norms.

**References**
1. Aydın, Ö., & Karaarslan, E. (2022). OpenAI ChatGPT generated literature review: Digital twin in healthcare. Available at SSRN 4308687.
2. Kosinski, M. (2023). Theory of mind may have spontaneously emerged in large language models. arXiv preprint arXiv:2302.02083.
3. Holterman, B., & van Deemter, K. (2023). Does ChatGPT have Theory of Mind?. arXiv preprint arXiv:2305.14020.
4. Rao, H., Leung, C., & Miao, C. (2023). Can chatgpt assess human personalities? a general evaluation framework. arXiv preprint arXiv:2303.01248.
5. Huang, J. T., Wang, W., Lam, M. H., Li, E. J., Jiao, W., & Lyu, M. R. (2023). ChatGPT an ENFJ, Bard an ISTJ: Empirical Study on Personalities of Large Language Models. arXiv preprint arXiv:2305.19926.
6. Nisbett, R. (2004). The geography of thought: How Asians and Westerners think differently... and why. Simon and Schuster.
7. Nisbett, R. E., Peng, K., Choi, I., & Norenzayan, A. (2001). Culture and systems of thought: holistic versus analytic cognition. Psychological review, 108(2), 291.
8. Nisbett, R. E., & Miyamoto, Y. (2005). The influence of culture: holistic versus analytic perception. Trends in cognitive sciences, 9(10), 467-473.
9. Norenzayan, A., Smith, E. E., Kim, B. J., & Nisbett, R. E. (2002). Cultural preferences for formal versus intuitive reasoning. Cognitive science, 26(5), 653-684.
10. Peng, K., Spencer-Rodgers, J., & Nian, Z. (2006). Naïve dialecticism and the Tao of Chinese thought. In Indigenous and cultural psychology: Understanding people in context (pp. 247-262). Boston, MA: Springer US.
11. Choi, I., Koo, M., & Choi, J. A. (2007). Individual differences in analytic versus holistic thinking. Personality and social psychology bulletin, 33(5), 691-705.
12. Talhelm, T., Zhang, X., Oishi, S., Shimin, C., Duan, D., Lan, X., & Kitayama, S. (2014). Large-scale psychological differences within China explained by rice versus wheat agriculture. Science, 344(6184), 603-608.
13. Dong, X., Talhelm, T., & Ren, X. (2019). Teens in rice county are more interdependent and think more holistically than nearby wheat county. Social Psychological and Personality Science, 10(7), 966-976.
14. Spencer-Rodgers, J., Boucher, H. C., Mori, S. C., Wang, L., & Peng, K. (2009). The dialectical self-concept: Contradiction, change, and holism in East Asian cultures. Personality and Social Psychology Bulletin, 35(1), 29-44.
15. Hamamura, T., Heine, S. J., & Paulhus, D. L. (2008). Cultural differences in response styles: The role of dialectical thinking. Personality and Individual Differences, 44(4), 932-942.
16. Singelis, T. M. (1994). The measurement of independent and interdependent self-construals. Personality and social psychology bulletin, 20(5), 580-591.

17. Krishna, A., Zhou, R., & Zhang, S. (2008). The effect of self-construal on spatial judgments. Journal of Consumer Research, 35(2), 337-348.
18. Kühnen, U., Hannover, B., & Schubert, B. (2001). The semantic–procedural interface model of the self: The role of self-knowledge for context-dependent versus context-independent modes of thinking. Journal of personality and social psychology, 80(3), 397.